\documentclass[10pt,journal,compsoc]{IEEEtran}
\ifCLASSOPTIONcompsoc
  \usepackage[nocompress]{cite}
\else
  \usepackage{cite}
\fi
\usepackage{amssymb}

\ifCLASSINFOpdf
   \usepackage[pdftex]{graphicx}
\else
\fi
\usepackage{amsmath}
\usepackage{gensymb}
\usepackage{colortbl}
\usepackage{svg}
\usepackage{multirow}
\usepackage{acronym}
\newacro{CNN}[CNN]{Convolutional Neural Network}
\newacro{BoCF}[BoCF]{Bag of Color Features}
\newacro{FC4}[FC$^4$]{Fully Convolutional Color Constancy}
\newacro{MCDE}[MCDE]{Monte Carlo Dropout Ensembles}
\newacro{RAE}[RAE]{recovery angular error}
\newacro{CWCC}[CWCC]{Channel-Wise Color Constancy}
\makeatletter
\renewcommand\@makefntext[1]{\leftskip=2em\hskip-2em\@makefnmark#1}
\makeatother

\begin{document}

\title{Learning to ignore: rethinking attention in CNNs}

\author{Firas~Laakom*, Kateryna~Chumachenko*,
Jenni~Raitoharju,
Alexandros~Iosifidis, and~Moncef~Gabbouj
       \thanks{* Equal contribution}
       \thanks{F. Laakom,   K. Chumachenko and M. Gabbouj are with Department of Computing Sciences, Tampere University, Tampere, Finland, Tampere University, Tampere, Finland.}
       \thanks{J. Raitoharju is with the Programme for Environmental Information, Finnish Environment Institute, Jyväskylä, Finland.}
       \thanks{A. Iosifidis is with the Department of Electrical and Computer Engineering, Aarhus University, Aarhus, Denmark.}

}

\IEEEtitleabstractindextext{%
\begin{abstract}
Recently, there has been an increasing interest in applying attention mechanisms in Convolutional Neural Networks (CNNs) to solve computer vision tasks. Most of these methods learn to explicitly identify and highlight relevant parts of the scene and pass the attended image to further layers of the network. In this paper, we argue that such an approach might not be optimal. Arguably, explicitly learning which parts of the image are relevant is typically harder than learning which parts of the image are less relevant and, thus, should be ignored.  In fact, in vision domain, there are many easy-to-identify patterns of irrelevant features. For example, image regions close to the borders are less likely to contain useful information for a classification task. Based on this idea, we propose to reformulate the attention mechanism in CNNs to learn to ignore instead of learning to attend. Specifically, we propose to explicitly learn irrelevant information in the scene and suppress it in the produced representation, keeping only important attributes. This implicit attention scheme can be incorporated into any existing attention mechanism. In this work, we validate this idea using two recent attention methods Squeeze and Excitation (SE) block and Convolutional Block Attention Module (CBAM). Experimental results on different datasets and model architectures show that learning to ignore, i.e., implicit attention, yields superior performance compared to the standard approaches.
\end{abstract}

}

\maketitle

%-------------------------------------------------------------------------
\IEEEraisesectionheading{\section{Introduction}\label{sec:intro}}
\IEEEPARstart{I}{nspired} by the properties of the human visual system, attention mechanisms have been recently applied in the field of deep learning,  resulting in improved performance of the existing models across multiple applications. In the context of computer vision, learning to attend, i.e., learning to highlight and emphasize relevant attributes of images, have led to development of novel approaches \cite{se, cbam} in Convolutional Neural Networks (CNNs), improving their capabilities in many tasks \cite{att1, jiang2018self, xu2015show}. 

Related to the concept of attention, recent studies in neuroscience suggest that the ability of humans to successfully perform visual tasks is related to the ability to ignore and suppress distractive information \cite{cosman2018prefrontal, ignore1, gaspelin2018role}. For example, the authors of \cite{ignore1} show that differences in visual working memory capacity, i.e., ability to remember visual features of multiple objects, are specifically related to distractor-suppression activity in visual cortex. This idea is reinforced in \cite{gaspelin2018role}, where the authors provide evidence on an inhibitory mechanism of suppression of salient distractors aimed at preventing them from capturing attention and being further processed by humans. Additional studies \cite{cunningham2016taming} report that ignoring the irrelevant information is a powerful learning tool for human cognition with ubiquitous effectiveness. Inspired by these findings, we investigate the intuition of learning to explicitly ignore irrelevant information in the field of computer vision and reformulate attention mechanisms commonly utilized in CNNs under the framework of learning to ignore rather than learning to attend.

Existing attention mechanisms used in CNNs learn the attention masks by directly optimizing for the high response of attributes of the image that are important for the prediction and, thus, should be focused on more. The learned attention masks are applied to feature representations, leading to higher emphasis put on the attributes of interest, and, therefore, resulting in implicit ignoration of the irrelevant features. In our work, we propose to rethink this logic and instead explicitly focus on ignoring irrelevant regions, hence achieving the attention to important regions implicitly. 
We argue that learning of features that should be ignored is an easier task than learning to attend and, therefore, optimization with such an objective leads to better training. Arguably, discriminative features of samples of different classes are harder to capture and often require more advanced feature learning. On the other hand, irrelevant attributes or attributes common between classes are often related to easy-to-identify patterns, such as borderline locations on the image or background features that can already be learned at early stages of training. 
Following this intuition, we design our method to explicitly optimize which attributes of the image should be ignored, and based on this, the important attributes that should be attended are derived implicitly. We validate this idea using two recent attention methods Squeeze and Excitation (SE) block and Convolutional Block Attention Module (CBAM) and show that indeed our intuition holds and explicitly learning features to ignore leads to better model performance.

Our contributions can be summarized as follows:
\begin{itemize}
    \item We propose a new perspective on attention in computer vision where the main aim is to learn to ignore instead of learning to attend. 
    \item We propose an implicit attention scheme which explicitly learns to identify the irrelevant parts of the scene and suppress them. The proposed approach can be incorporated into any existing attention mechanism.
    \item We validate this idea using two attention mechanisms. Specifically, we reformulate Squeeze-and-Excitation (SE) block and Convolutional Block Attention Module (CBAM) using our paradigm, i.e., \textit{learn to ignore}, and show the superiority of such an approach.
\end{itemize}
\section{Related work}
\textbf{Attention mechanisms in vision.} The idea of attention in vision tasks stems from the properties of selective focus in the human visual system, i.e., that humans do not perceive images as a whole, but rely on certain salient parts of them. This property gave rise to a variety of attention-based learning mechanisms aimed to enhance the performance in computer vision domain \cite{att1,jiang2018self, att2}, finding its applications in a variety of tasks, including sequence learning \cite{vaswani2017attention}, image captioning \cite{xu2015show}, and others \cite{wu2019spatio, zhang2019relation}. A subset of attention-driven methods is directed at CNNs and aims at selecting and highlighting relevant attributes in the feature space during training \cite{se, cbam}. Conventionally, this is achieved by learning attention masks over feature representations that encode the importance of different attributes in form of weights and applying these masks on intermediate feature representations. This results in higher influence of features relevant for decision making in subsequent layers. 

Other tasks adjacent to this line of research include saliency estimation, image segmentation, and weakly-supervised object localization. In saliency estimation, the goal is to estimate salient, i.e., significant regions of the scene without any prior knowledge on the scene in unsupervised ~\cite{aytekin2018probabilistic,zhang2018deep} or supervised manner~\cite{liu2018picanet,liu2021visual,liu2016dhsnet}. In image segmentation, the task is to partition a given image into a set of segments, based on either semantics (semantic segmentation) or individual objects (instance segmentation) \cite{minaee2021image}. In weakly-supervised object localization, the goal is to predict the location of the object given only image-level labels \cite{zhang2021weakly}.

Within the attention mechanisms utilized in CNNs, two of the notable ones include Squeeze-and-Excitation block  (SE) \cite{se} and Convolutional Block Attention Module (CBAM) \cite{cbam}. In SE, an attention mask is learned channel-wise based on global average-pooled features of intermediate representations and applied at multiple layers of the ResNet architecture \cite{he2016deep}. A further extension is the CBAM method that enriches the SE mechanism by additional max-pooled input and learns spatial attention in addition to channel-wise one. The learned attention weight masks are then applied channel-wise or pixel-wise to corresponding feature maps. These methods were shown to lead to superior performance across various domains and can be incorporated in any CNN architecture.

\textbf{Learning by ignoring.} 
Learning by ignoring is a powerful learning paradigm, which has been used in various machine learning applications \cite{gu2019improved,jiang2013saliency,zhao2020learning}. It has been leveraged in the context of saliency estimation \cite{li2013saliency,jiang2013saliency,zhu2014saliency,aytekin2018probabilistic}.  For example, the authors of \cite{aytekin2018probabilistic} propose an unsupervised graph-based saliency estimation approach, where  auxiliary variables are used to encode prior knowledge on regions to be ignored, such as dark regions, as it is assumed that they are less-likely to contain salient object. A similar approach was proposed for the color constancy problem \cite{laakom2020probabilistic}.  In the context of machine translation, it has been shown that learning to ignore spurious correlations in the data can improve the performance of neural networks in zero-shot translation \cite{gu2019improved}.  In the context of domain adaptation, a learning framework assigning and learning an 'ignoring' score for each training sample and re-weighting the total loss based on these scores was proposed in \cite{zhao2020learning}.

\section{Learning to ignore in CNNs} \label{learn_ignore}
Attention in CNNs is generally formulated in a form of a learned attention mask that emphasizes relevant information in a feature map. Formally, given a feature map $\mathbf{F}$, attention can be defined as follows:
\begin{equation} \label{explicit_attention}
  \mathbf{F}' = \mathbf{F} \otimes f_\theta(\mathbf{F}),
\end{equation}
where  $\mathbf{F}'$ is the attended feature map output, $\otimes$ is the element-wise multiplication and $f_\theta(\cdot)$ is an attention function with learnable parameters $\theta$, which takes as input a feature map $\mathbf{F}$ and returns an attention mask $f_\theta(\mathbf{F}) \in [0,1]$. This mask is then element-wise multiplied with the original map $\mathbf{F}$ in order to produce the output map $\mathbf{F}'$.  The mask $f_\theta(\mathbf{F})$ is expected to identify relevant spatial or channel information and output the 'importance score' for each attribute, producing high response for most relevant regions and smaller values for regions of lesser interest.  This can be seen as an explicit attention mechanism, where the model  $f_\theta(\cdot)$  learns to directly identify and highlight relevant information. 

In this work, we develop a new formulation of the concept of attention in CNNs, 
where the main target is learning to ignore instead of learning to attend. By training the model to predict irrelevance of features, rather than their importance, we expect to simplify the training objective and, hence, to improve the learning of the model. Our approach consists of a function which learns to identify irrelevant or confusing parts of the feature map in order to suppress them, followed by inversion of predicted irrelevance scores. Formally, this can be formulated as follows:
\begin{equation} \label{implicit_attention}
  \mathbf{F}' = \mathbf{F} \otimes T(g_\theta(\mathbf{F})),
\end{equation}
where  $g_\theta(\cdot)$ is a function with learned parameters $\theta$ that is expected to learn to highlight information in the feature map that is irrelevant or confusing for the prediction. This can be seen as an \textit{ignoring mask} that outputs high values for attributes and regions that should be suppressed in the feature map. The function $T(\cdot)$ is a function with an output $T(x)$ inversely proportional to $x$, hence flipping the learned ignoring mask and transforming it into an attention mask. Similarly to Eq.~\eqref{explicit_attention}, the final feature map $\mathbf{F}'$ is obtained by element-wise multiplication of the input map $\mathbf{F}$ and the flipped ignoring mask $T(g_\theta(\mathbf{F}))$.

Given an ignoring mask $g_\theta(\mathbf{F})$, the function $T(\cdot)$ can be any function satisfying the condition of being inversely proportional to its input and bounded between $[0,1]$. In this work, we propose three variants:

\begin{equation} 
T_1(x)= 1 - \alpha x,
\end{equation}
\begin{equation}T_2(x)= sigmoid(\frac{1}{x}),\end{equation} 
\begin{equation}T_3(x)= sigmoid(-x).\end{equation}
The first variant $T_1(\cdot)$ linearly converts the ignoring mask to an attention one, and $\alpha$ is a hyper-parameter controlling  this linear scaling.  The extreme case $\alpha=0$ corresponds to the extreme case $\mathbf{F}' = \mathbf{F}$, i.e., none of the features are emphasized or suppressed. For the second and third variants $T_2$ and $T_3$, a sigmoid function is applied to ensure that the output is bounded between $[0,1]$.

We argue that formulating the objective as learning of irrelevant features that should be ignored, as opposed to focusing on important features, is beneficial, as optimization of a model with such an objective is easier. This is due to potential presence of many easy-to-identify patterns of irrelevant attributes, such as borderline pixel locations, color and lighting perturbations, or background properties that are not correlated with the groundtruth labels. At the same time, information responsible for predictions is generally label-specific and harder to capture. Moreover, learning of discriminative attributes that can  be regarded as important often requires learning of complex feature representations that can be achieved only at latter stages of training, while patterns irrelevant for decision making can often be identified already at the early stages.

It can be argued that standard attention, i.e., Eq.~\eqref{explicit_attention}, is also learning to ignore as it is expected to indirectly assign smaller values for less important regions. 
However, function $f_\theta(\cdot)$ is optimized directly for highlighting relevant information and, hence, this can be seen as an implicit and indirect strategy of learning to ignore.
In our approach, Eq.~\eqref{implicit_attention}, the model $g_\theta(\cdot)$ is explicitly optimized for identifying the irrelevant or confusing parts and the function $T(\cdot)$ suppresses them. This can be seen as an implicit learning to attend approach and explicit learning to ignore approach, as opposed to the standard attention which has an explicit learning to attend formulation.  

As can be seen, the main difference between implicit and explicit attention formulations is the presence of a flipping function $T(\cdot)$. It can be seen from Eq.~\eqref{explicit_attention} and Eq.~\eqref{implicit_attention} that $f_\theta(\cdot)$ can be directly replaced by $T(g_\theta(\cdot))$. This makes it straightforward to reformulate any existing explicit attention method to learn to ignore instead of learning to attend by applying an inversion function $T(\cdot)$ on top of the learned mask. This way, the model $g_\theta(\cdot)$ can be learned as the model $f_\theta(\cdot)$ in conventional attention methods, while its parameters will be optimized to detect irrelevant or confusing regions instead of relevant ones. In this paper, for the choice of the function $f_\theta(\cdot)$, we consider two state-of-the-art attention mechanisms, namely  SE \cite{se} and CBAM \cite{cbam} , and we show how to reformulate them using our paradigm in the following subsections.

\subsection{Ignoring with Squeeze-and-Excitation blocks}

Squeeze-and-Excitation (SE) block \cite{se} presents a mechanism to learn channel-wise attention, focusing on which features of the representation are important for prediction. This is achieved by squeezing the spatial information into a channel representation, followed by an excitation operation that highlights important channels via a bottleneck block. Formally, given a feature map $\mathbf{F}$, this is defined as follows:

\begin{equation}
    f_\theta(\mathbf{F}) = \sigma(\mathbf{W}_2\delta(\mathbf{W}_1GAP(\mathbf{F}))),
\end{equation}
where $GAP(\cdot)$ denotes Global Average Pooling, $\delta$ is a ReLU activation, $\sigma$ is the sigmoid function, $\mathbf{W}_1 \in \mathbb{R}^{c\times\frac{c}{r}}$ and $\mathbf{W}_2 \in \mathbb{R}^{\frac{c}{r}\times c}$ are linear layers, $c$ is the number of channels in $\mathbf{F}$,  and $r$ is the reduction rate in the bottleneck block. Given the output $f_\theta(\mathbf{F})$, the attended feature map is obtained by applying the learned mask element-wise between corresponding channels.

To incorporate our ignoring paradigm into SE, we apply $T(\cdot)$ to the output $f_\theta(\mathbf{F})$, hence transforming its objective into learning features that should be ignored. Specifically, we define the three variants as: $f^1_\theta(\mathbf{F}) = 1 - \alpha \sigma(\mathbf{W}_2\delta(\mathbf{W}_1GAP(\mathbf{F})))$; $f^2_\theta(\mathbf{F}) = \sigma(\frac{1}{\sigma(\mathbf{W}_2\delta(\mathbf{W}_1GAP(\mathbf{F})))})$; $f^3_\theta(\mathbf{F}) = \sigma(-\mathbf{W}_2\delta(\mathbf{W}_1GAP(\mathbf{F})))$ using the definitions of $T_1$, $T_2$, and $T_3$, respectively. As can be noticed, in the first two variants $T(\cdot)$ is applied directly on $f_\theta(\mathbf{F})$, while in the third case it is applied on pre-sigmoid output to ensure sufficiently wide range for attention scores.

\subsection{Ignoring with Convolutional Block Attention Modules}

Following the approach of SE, Convolutional Block Attention Module (CBAM) \cite{cbam} extends it to incorporate spatial attention  as well as to enrich channel attention with an additional input representation. Under the definition of attention in Eq.~\eqref{explicit_attention}, this is formulated as follows:
\begin{equation}
\begin{gathered}
    f^{ch}(\mathbf{F}) = \sigma(\mathbf{W}_2\delta(\mathbf{W}_1(GAP(\mathbf{F}))) + \mathbf{W}_2\delta(\mathbf{W}_1(GMP(\mathbf{F})))), \\
    \mathbf{F}^{ch} = \mathbf{F} \otimes  f^{ch}(\mathbf{F}), \\
 f^{sp}(\mathbf{F}^{ch}) = \sigma(Conv^{7 \times 7}(GAP(\mathbf{F}^{ch}) \frown GMP(\mathbf{F}^{ch}))),
    \end{gathered}
\end{equation}
where $f^{ch}$ and $f^{sp}$ denote channel and spatial attention, respectively, $GAP(\cdot)$ and $GMP(\cdot)$ correspond to Global Average Pooling and Global Max Pooling, respectively, $\delta$ is a ReLU activation, $\sigma$ is the sigmoid activation, $\mathbf{W}_1 \in \mathbb{R}^{c\times\frac{c}{r}}$ and $\mathbf{W}_2 \in \mathbb{R}^{\frac{c}{r}\times c}$ are linear layers, $c$ is the number of channels in $\mathbf{F}$,  and $r$ is the reduction rate in the bottleneck block, similarly to SE. $\mathbf{F}^{ch}$ is the channel-wise attended feature map, $Conv^{7\times7}$ denotes a convolutional layer with $7 \times 7 $ kernel, and $\frown$ denotes concatenation. 

As can be seen, channel and spatial attention masks are applied sequentially and channel-attended feature representations are used as input to compute spatial attention. Following this, we transform CBAM for ignoring by addition of inversion function $T(\cdot)$ on top of both channel function $f^{ch}(\cdot)$ and spatial function $f^{sp}(\cdot)$ to reformulate their objectives as learning of features and regions to ignore. In both cases, variants of $T_1(\cdot)$ and $T_2(\cdot)$ are applied directly on the output of corresponding functions, and $T_3(\cdot)$ is applied on pre-sigmoid output.

\section{Experimental Results}

\subsection{CIFAR10 \& CIFAR100}

We start by validating our approach on image classification task using CIFAR10 and CIFAR100 \cite{cifar} datasets. To show invariance of the proposed approach to specific model architectures, we evaluate two state-of-the-art CNNs, namely, ResNet50 \cite{he2016deep} and DenseNet \cite{densenet} architectures. We report the results of standard models with no attention, models with applied CBAM and SE attention blocks, and models with our proposed ignoring approach with both CBAM and SE variants with the three inversion function variants presented in Section~\ref{learn_ignore}.

All the models are optimized using Stochastic Gradient Descent (SGD) \cite{ruder2016overview} with a momentum of 0.9 \cite{rumelhart1986learning}, weight decay of $0.0001$ \cite{krogh1992simple}, and a batch size of 128. The initial learning rate is set to 0.1 and is then decreased by a factor of 5 after 60, 120, and 160 epochs, respectively. The models are trained for 200 epochs  with the best performance on the validation set used for testing. Each experiment is repeated three times and the average performance is reported. 40k images are used for training and 10k for validation. Standard data augmentation is used \cite{huang2017densely,zhang2018mixup}.

\begin{table*}[h]
\centering
\begin{tabular}{|cl|l|l|l|}

\hline
\multicolumn{2}{|l|}{}    & \multicolumn{1}{|c|}{CIFAR 10}  & \multicolumn{2}{|c|}{CIFAR 100}  \\
 \hline
         &                    & Top-1 Error\%                             & Top-1 Error\%                            & Top-5 Error\%                     \\ \hline \hline
\multicolumn{1}{|l|}{\multirow{7}{*}{\rotatebox{90}{ResNet50}}} & Standard            & 08.27 $\pm$ 0.54            & 34.06 $\pm$ 1.02          & 10.97 $\pm$ 0.54  \\  
  \multicolumn{1}{|l|}{}       & SE                 & 07.63 $\pm$ 0.37            & 32.80 $\pm$ 0.11          & 09.97 $\pm$ 0.50   \\
 \multicolumn{1}{|l|}{}        & SE-Ign$_{1(\alpha=1)} $          & 07.42 $\pm$ 0.29 & 32.50 $\pm$ 0.26 & 09.92 $\pm$ 0.37  \\
 \multicolumn{1}{|l|}{}        & SE-Ign$_{1(\alpha=0.5)} $       & 07.61 $\pm$ 0.46 & 31.40 $\pm$ 0.68       & \textbf{09.39 $\pm$ 0.19 }  \\
 \multicolumn{1}{|l|}{}        & SE-Ign$_{1(\alpha=0.8)} $       & 07.76 $\pm$ 0.73             & 32.71 $\pm$ 1.15           & 10.07 $\pm$ 0.64   \\
 \multicolumn{1}{|l|}{}        & SE-Ign$_2$    & 07.66 $\pm$ 0.13       & 32.78 $\pm$ 0.77           & 10.11 $\pm$ 0.56    \\
  \multicolumn{1}{|l|}{}       & SE-Ign$_3$      & \textbf{07.28 $\pm$ 0.17} & \textbf{30.95 $\pm$ 0.08}           & 09.49 $\pm$ 0.36    \\ \hline
\multicolumn{1}{|l|}{\multirow{7}{*}{\rotatebox{90}{DenseNet}}} & Standard            & 07.07 $\pm$ 0.33           & 29.25 $\pm$ 0.10           & 08.26 $\pm$ 0.12      \\
 \multicolumn{1}{|l|}{}        & SE                 & 06.96 $\pm$ 0.05            & 29.43 $\pm$ 0.44           & 08.36 $\pm$ 0.33  \\
  \multicolumn{1}{|l|}{}       & SE-Ign$_{1(\alpha=1)}$           & 06.94 $\pm$ 0.07         & 29.17 $\pm$ 0.07          & 08.22 $\pm$ 0.13  \\
  \multicolumn{1}{|l|}{}       & SE-Ign$_{1(\alpha=0.5)}$       & 06.69 $\pm$ 0.04                     & \textbf{27.64 $\pm$ 0.30}           & \textbf{07.30 $\pm$ 0.10}       \\
 \multicolumn{1}{|l|}{}        & SE-Ign$_{1(\alpha=0.8)}$        & 06.95 $\pm$ 0.14         & 27.73 $\pm$ 0.41          & 07.39 $\pm$ 0.07  \\
\multicolumn{1}{|l|}{}         & SE-Ign$_2$ & 06.80 $\pm$ 0.09           & 28.08 $\pm$ 0.35         & 07.39 $\pm$ 0.23     \\
\multicolumn{1}{|l|}{}         & SE-Ign$_3$  & \textbf{06.41 $\pm$ 0.08}         & 27.77 $\pm$ 0.54          & 07.65 $\pm$ 0.20   \\
\hline
\end{tabular}
\caption{Results of SE variants on CIFAR10 and CIFAR100 datasets.}
 \label{SE_cifar}

\end{table*}

In Table \ref{SE_cifar}, we report the experimental results of the standard model, i.e., no attention, SE, and our different SE-based variants, namely, SE-Ign$_i$ where i indicates the flipping function used ($T_1$ or $T_2$ or $T_3$). For the first variant, i.e., SE-Ign$_1$, we experiment with three different values of hyper-parameter $\alpha$: 1, 0.8, and 0.5. We note that for both architectures applying  an explicit or implicit attention mechanism  consistently  outperforms the standard model. On CIFAR10, the best performance is achieved using our third variant, i.e., SE-Ign$_3$, which improves the results by 1\% compared to standard and +0.3\% compared SE using ResNet50 architecture. On CIFAR100, the lowest top1-\% error rates are achieved by SE-Ign$_3$ and SE-Ign$_{1(\alpha=0.5)}$ for ResNet50 and DenseNet architectures, respectively. In fact, on this dataset our third variant boosts the accuracy by 4\% compared to the standard and 1.85\% compared to SE. This can be explained by the fact that for this dataset only 500 training samples per class are available, thus making it hard to directly learn the relevant visual features for each class. At the same time, the irrelevant features are more universal and typically independent of the class, thus making them easier to learn in a scarce data context. 

In Table \ref{CBAM_cifar}, we report the empirical results for the different CBAM-based variants. As can be seen, the results with this attention variant are consistent with our findings using SE. For both datasets and for both architectures, learning to ignore  yields better performance compared to both the standard model and the SE attention.  The top performance is achieved by either by CBAM-Ign$_{1(\alpha=0.5)}$ or  CBAM-Ign$_{1(\alpha=0.8)}$ variant.  More results can be found Supplementary material Table 1.

\begin{table*}[h]
\centering
\begin{tabular}{|cl|l|l|l|}

\hline
\multicolumn{2}{|l|}{}    & \multicolumn{1}{|c|}{CIFAR 10}  & \multicolumn{2}{|c|}{CIFAR 100}  \\
 \hline
         &                      & Top-1 Error\%                        & Top-1 Error\%                          & Top-5 Error\%                     \\ \hline \hline
\multicolumn{1}{|l|}{\multirow{7}{*}{\rotatebox{90}{ResNet50}}} & Standard              & 08.27 $\pm$ 0.54      & 34.06 $\pm$ 1.02 & 10.97 $\pm$ 0.54    \\
\multicolumn{1}{|l|}{}          & CBAM                 & 08.04 $\pm$ 0.03 & 31.46 $\pm$ 0.20   & 09.32 $\pm$ 0.15     \\
\multicolumn{1}{|l|}{}          & CBAM-Ign$_{1(\alpha=1)}$            & 07.78 $\pm$ 0.28        & 31.03 $\pm$ 0.25       & 09.28 $\pm$ 0.27    \\
 \multicolumn{1}{|l|}{}         & CBAM-Ign$_{1(\alpha=0.5)}$        &\textbf{07.17 $\pm$ 0.05}      & 30.58 $\pm$ 0.20        & 09.25 $\pm$ 0.23   \\
\multicolumn{1}{|l|}{}          & CBAM-Ign$_{1(\alpha=0.8)}$        & 07.40 $\pm$ 0.23                   & \textbf{30.28 $\pm$ 0.39 }       & \textbf{09.08 $\pm$ 0.33 }   \\
\multicolumn{1}{|l|}{}          & CBAM-Ign$_2$    & 07.53 $\pm$ 0.29       & 31.42 $\pm$ 0.58       & 09.27 $\pm$ 0.21    \\
\multicolumn{1}{|l|}{}          & CBAM-Ign$_3$      & 07.60 $\pm$ 0.10        & 30.88 $\pm$ 0.22        & 09.38 $\pm$ 0.32    \\
         \hline
\multicolumn{1}{|l|}{\multirow{7}{*}{\rotatebox{90}{DenseNet}}} & Standard              & 07.07 $\pm$ 0.33     & 29.25 $\pm$ 0.10         & 08.26 $\pm$ 0.12      \\
\multicolumn{1}{|l|}{}          & CBAM                 & 07.21 $\pm$ 0.23      & 30.63 $\pm$ 0.23         & 08.90 $\pm$ 0.14   \\
 \multicolumn{1}{|l|}{}         & CBAM-Ign$_{1(\alpha=1)}$           & 07.19 $\pm$ 0.26     & 29.63 $\pm$ 0.46        & 08.37 $\pm$ 0.39  \\
 \multicolumn{1}{|l|}{}         & CBAM-Ign$_{1(\alpha=0.5)}$       & 06.53 $\pm$  0.14               & 27.92 $\pm$ 0.19        & 07.58 $\pm$ 0.27    \\
 \multicolumn{1}{|l|}{}         & CBAM-Ign$_{1(\alpha=0.8)}$        & \textbf{06.40 $\pm$ 0.14}      & \textbf{27.11 $\pm$ 0.08}        & \textbf{07.33 $\pm$ 0.19}     \\
 \multicolumn{1}{|l|}{}         & CBAM-Ign$_2$ & 06.80 $\pm$ 0.02     & 27.88 $\pm$ 0.59        & 07.62 $\pm$ 0.05     \\
 \multicolumn{1}{|l|}{}         & CBAM-Ign$_3$   & 06.68 $\pm$ 0.05       & 27.94 $\pm$ 0.10        & 07.78 $\pm$ 0.21   \\
 \hline
\end{tabular}
\caption{Results of CBAM variants on CIFAR10 and CIFAR100 datasets. }
 \label{CBAM_cifar}
\end{table*}

\subsection{ImageNet}
To further validate the effectiveness of our learning to ignore framework, we perform additional experiments on ImageNet dataset \cite{deng2009imagenet} using ResNet50. For training on ImageNet, optimization is done with SGD with the same weight decay and momentum as used for CIFAR datasets. The initial learning rate is set to 0.1 and reduced by a factor of 10 after 30, 60, and 80 epochs, respectively. The models are trained for 90 epochs with batch size of 256 with the results reported on the validation set. 

\begin{table}[]
\centering
\begin{tabular}{|l|c|c|}
\hline
                                                & Top-1 Error\% & Top-5 Error\%  \\ \hline                                                \hline

 Standard  & 23.73   & 06.85 \\
\hline
SE     & 22.70 & 06.35 \\  
SE-Ign$_{1(\alpha=1)}$ &  22.60  &  \textbf{06.29} \\
SE-Ign$_{1(\alpha=0.5)}$ & 23.03   & 06.58  \\ 
SE-Ign$_{1(\alpha=0.8)}$ &  22.88  & 06.30  \\ 
SE-Ign$_2$ &  23.16  & 06.55  \\ 
SE-Ign$_3$ &  \textbf{22.59}  & 06.32  \\ 
\hline
CBAM     & 22.91 & 06.58 \\  
CBAM-Ign$_{1(\alpha=1)}$ &   \textbf{22.84}  &  06.50\\
CBAM-Ign$_{1(\alpha=0.5)}$ &  \textbf{22.84}  &  06.52 \\
CBAM-Ign$_{1(\alpha=0.8)}$ & \textbf{22.84}   &  06.40 \\
CBAM-Ign$_2$ & 23.02 &  \textbf{06.39} \\
CBAM-Ign$_3$ &  23.10 & 06.44  \\
\hline
\end{tabular}
\caption{Results of CBAM and SE with variants of ignoring on ImageNet dataset}
\end{table}

Table 3 shows the results on ImageNet dataset, where Top-1 and Top-5 errors are reported. As can be seen, our results are consistent with findings on CIFAR10 and CIFAR100 datasets. Specifically, we find that applying attention, whether explicit or implicit, outperforms standard model. At the same time, the proposed framework based on ignoring outperforms the conventional attention in a vast majority of cases. In SE variant, SE-Ign$_{1(\alpha=1)}$ and SE-Ign$_3$ outperform the conventional approach, while other variants report competitive results with minimal gap. Best result of SE-Ign$_3$ outperforms the standard model by $1.1\%$. In CBAM, all variants of CBAM-Ign$_1$ outperform conventional approach on both Top-1 and Top-5 metric, and CBAM-Ign$_2$ and CBAM-Ign$_3$ outperform conventional CBAM on Top-5 metric, while being competitive on Top-1 metric. More results can be found Supplementary material Table 2.

\subsection{NTU-RGBD}

To further demonstrate the effectiveness of our approach, we additionally evaluate the proposed method in the multimodal fusion setting. Here, we rely on the Multimodal Transfer Module (MMTM) \cite{mmtm} architecture for our evaluation. MMTM is a method for fusing information from multiple modalities in multiple-stream architectures, which has recently shown good performance in a variety of tasks, including activity recognition, gesture recognition, and audiovisual speech enhancement. 

The method relies on an architecture inspired from Squeeze-and-Excitation blocks placed between network branches. Specifically, considering a two-stream scenario, intermediate feature representations from two network branches corresponding to two modalities are first spatially squeezed into channel descriptors by applying global average pooling in each branch. The squeezed representations are subsequently concatenated and projected into a joint lower-dimensional space. The resulting features are transformed with two projection matrices corresponding to each of the two modalities to the spaces of original dimensionalities, and sigmoid activation is then applied to obtain attention masks. The masks are further multiplied element-wise with original feature representations in each branch. 

As can be seen, the fusion module is essentially a multi-modal SE-block with joint squeeze and modality-specific excitation operations, to which we apply our ignoring framework as described in Section~3.1. We perform experiments on NTU-RGBD dataset \cite{nturgbd} for human action recognition, where we fuse the skeleton and RGB modalities, similarly to MMTM \cite{mmtm}. We follow our ignoring paradigm and replace the SE attention mask in each branch with our proposed approach. The rest of the architecture and training protocol follows that of MMTM. We  initialize the model from ImageNet+Kinectics pretrained weights, finetune for 10 epochs with batch size 8, and report the test set performance of the model that performed best on validation set. The results are reported in Table~\ref{tab:mmtm}. As can be seen, the proposed ignoring approaches outperform the baseline in the vast majority of cases.

\begin{table*}[]
\centering
\begin{tabular}{|l|cccccc|}
\hline
    & MMTM  & Ign$_{1(\alpha=1)}$ & Ign$_{1(\alpha=0.5)}$ & Ign$_{1(\alpha=0.8)}$ & Ign$_2$ & Ign$_3$ \\
    \hline
NTU-RGBD & 89.98 & 89.99  & \textbf{90.52} & 88.70  & 90.21         & 90.36    \\
\hline
\end{tabular}
\caption{Accuracy on NTU-RGBD dataset}
\label{tab:mmtm}
\end{table*}

\subsection{Discussion}

\begin{figure*}[h]
\includegraphics[width=0.8\textwidth]{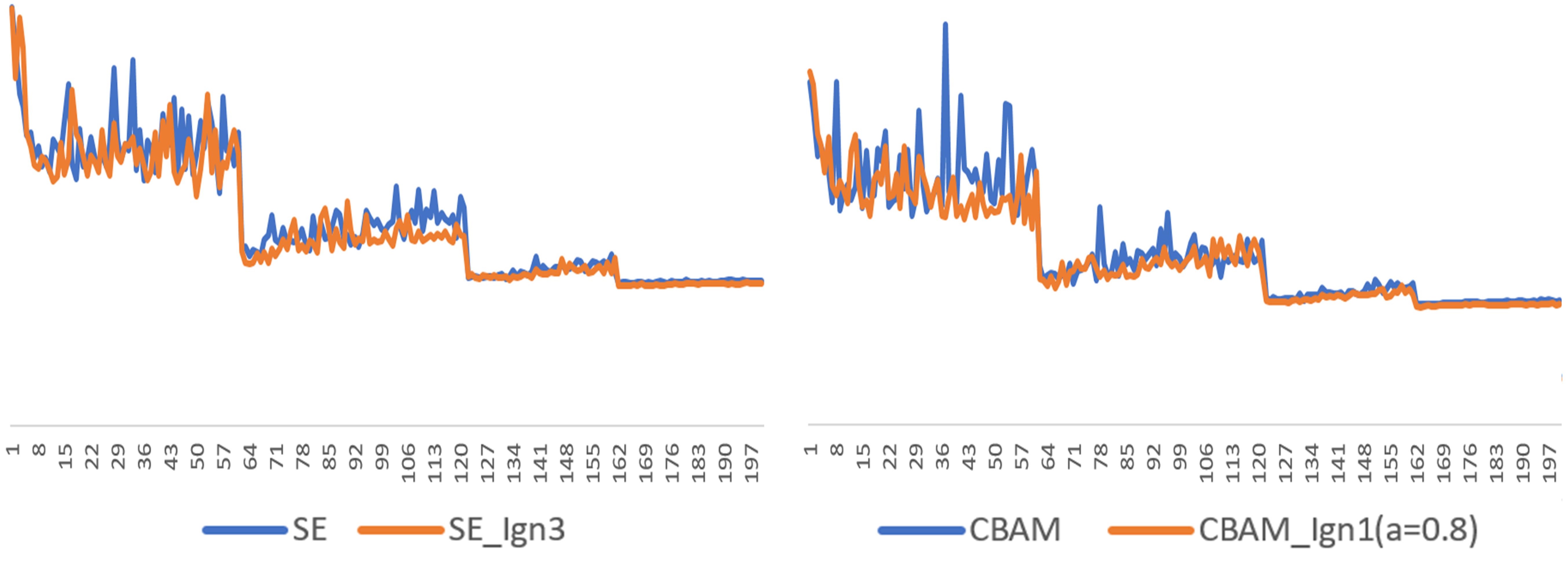}
\centering
 \caption{Validation loss curves of ResNet50 on CIFAR100 using the different attention approaches.}
 \label{train_curves}

\end{figure*}

As can be seen from the experimental results in previous sections, learning to ignore consistently yields superior performance compared to the baselines. We argue that this stems from the fact that learning irrelevant information is easier than identifying what should be attended. For example, in order to learn features that should be attended to, the model needs to first learn to extract patterns such as lines and edges and make associations with the class labels in order to produce a meaningful attention mask. On the other hand, irrelevant patterns, such as background textures and borderline pixels, are often shared across the dataset, are persistent and independent of the class labels, which makes them easier to learn. Therefore, it should be possible to learn them already in the early stages of training. Figure~\ref{train_curves} shows the validation loss curves of the baseline attention methods and the best-performing ignoring methods with ResNet50 on CIFAR100 dataset (more training curves can be found in supplementary material). As can be seen, especially at the earlier stages of training, our approach results in lower loss with less fluctuations and more stable training, hence supporting our claim. 
From an optimization point of view, in the case of $\alpha$=1, only the gradient of the attention blocks are flipped, and thus in the back-propagation, when they are summed with the gradient of the main block (which are not flipped), the total feedback carried to the earlier layers is different and does not correspond to a flipped version of the total sum of the standard attention. Thus, this yields different feedback and leads to a different optimal solution in the end of the training (Figure 7 in supplementary material).

Moreover, in  Figure~\ref{figure_attention}, we provide visual results of the class activation maps \cite{selvaraju2017grad} produced by the different models on three different samples from validation set of ImageNet. As can be seen, the learning to ignore formulation leads to different attention maps compared to the explicit attention, i.e., learning to attend. Noticeably, standard CBAM attention tries to capture the relevant parts of the image directly, leading to the prediction being made based on the small part of the input that is considered by the model as the most important. This leads to the possibility that the model can miss some important parts of the class of interest on the image. As an example, only one of the plants on the lower figure is considered in CBAM model, as well as only a side of the bus in the middle image. On the other hand, our approach by learning to identify the non-relevant background regions first and subsequently suppressing them, simplifies the problem and typically results in an attention mask that is broader and captures the object of interest better, hence reducing the risk of suppressing relevant attributes of it.

%This can be seen especially well in maps corresponding to CBAM-Ign$_1(\alpha=0.5)$ and CBAM-Ign$_3$.

\begin{figure*}[h]
 
\includegraphics[width=0.9\textwidth]{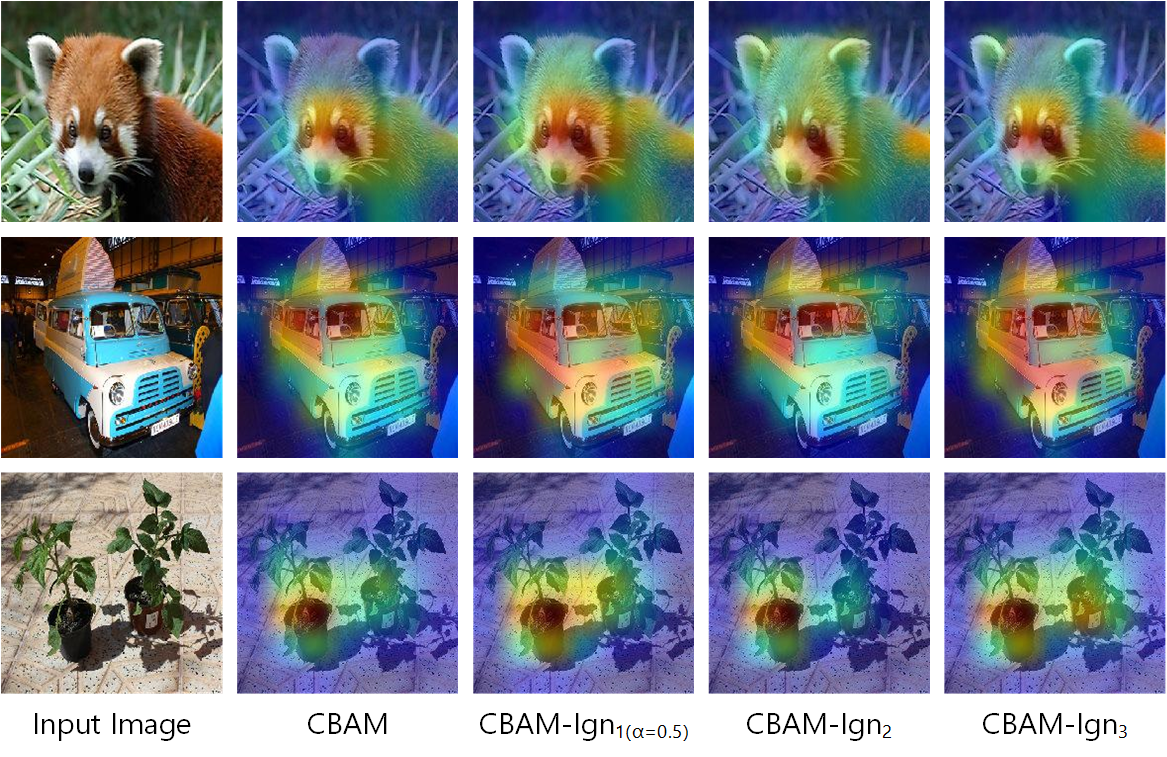}
\centering
 \caption{Visual results of different CBAM-based attention mechanisms on three different samples from validation set of ImageNet. The attention masks are obtained as in \cite{selvaraju2017grad}. 
 \label{figure_attention}}
\end{figure*}
\section{Conclusion}
In this paper, we provide a new perspective on attention in CNNs where the main target is learning to ignore instead of learning to attend. To this end, we propose an implicit attention scheme with three variants which can be incorporated into any existing attention mechanism. The proposed approach explicitly learns to identify  the irrelevant and confusing parts of the scene and suppresses them. In addition, we reformulate two state-of-the-art attention approaches, namely SE and CBAM, using our learning paradigm. Experimental results on three image classification datasets show that learning to ignore, i.e., implicit attention consistently outperforms standard attention across multiple models.

\ifCLASSOPTIONcompsoc
  \section*{Acknowledgments}
\else
  \section*{Acknowledgment}
\fi

This work has received funding from the European
Union’s Horizon 2020 research and innovation programme
under grant agreement No 871449 (OpenDR), and the NSF-Business  Finland
Center for Visual and Decision Informatics (CVDI) project AMALIA. 
The authors wish to acknowledge CSC – IT Center for
Science, Finland, for computational resources.

\ifCLASSOPTIONcaptionsoff
  \newpage
\fi

\bibliographystyle{IEEEtran}

\bibliography{strings}

\end{document}